\def\year{2021}\relax
\documentclass[letterpaper]{article} 
\usepackage{aaai21}  
\usepackage{times}  
\usepackage{helvet} 
\usepackage{courier}  
\usepackage[hyphens]{url}  
\usepackage{graphicx} 
\usepackage{natbib}  
\usepackage{caption} 
\usepackage{amsfonts}       
\usepackage{nicefrac}       
\usepackage{microtype}      
\usepackage{amsmath}
\usepackage{multicol}
\usepackage[displaymath, mathlines]{lineno}
\usepackage{subfigure}
\usepackage{commath}
\usepackage{amssymb}
\usepackage{nccmath}
\usepackage{float}
\usepackage{bm}
\usepackage{dsfont}
\usepackage{mdframed}
\usepackage{multirow}
\usepackage{tabularx}
\usepackage{tabulary}
\usepackage{amsthm}

\usepackage{tikz}
\usepackage{rotating}
\usepackage{mathtools}
\usepackage{booktabs}
\usepackage{adjustbox}
\usepackage{algorithm}
\usepackage{algpseudocode}
\usepackage{tabularx}
\usepackage[vlines]{tabularht}
\usepackage{subfiles} 
\usepackage{lineno}
\usepackage{wrapfig}

\iftrue
\usepackage{lastpage}
\usepackage{fancyhdr}
\fi
\usepackage{hyperref}

\title{CoxNTF: A New Approach for Joint Clustering and Prediction in Survival Analysis}

\author{
    Paul Fogel,\textsuperscript{\rm 1} Christophe Geissler,\textsuperscript{\rm 1} George Luta\textsuperscript{\rm 2, \thanks{To whom the correspondence should be addressed.}}\\
}
\affiliations{
    \small
    \textsuperscript{\rm 1}Data Services, Forvis Mazars, Levallois, France\\
    \textsuperscript{\rm 2} Department of Biostatistics, Bioinformatics and Biomathematics, Georgetown University Medical Center, Washington, DC, USA \\
    paul.fogel@forvismazars.com,
    christophe.geissler@forvismazars.com,
    george.luta@georgetown.edu
}

\begin{document}

\maketitle
\begin{abstract}
The interpretation of the results of survival analysis often benefits from latent factor representations of baseline covariates. However, existing methods, such as Nonnegative Matrix Factorization (NMF), do not incorporate survival information, limiting their predictive power. We present CoxNTF, a novel approach that uses non-negative tensor factorization (NTF) to derive meaningful latent representations that are closely associated with survival outcomes. CoxNTF constructs a weighted covariate tensor in which survival probabilities derived from the Coxnet model are used to guide the tensorization process. Our results show that CoxNTF achieves survival prediction performance comparable to using Coxnet with the original covariates, while providing a structured and interpretable clustering framework. In addition, the new approach effectively handles feature redundancy, making it a powerful tool for joint clustering and prediction in survival analysis. \newline

\textbf{Keywords}: NMF, NTF, Coxnet, latent representation
\end{abstract}

\section{Introduction}
Nonnegative Matrix Factorization (NMF) \cite{Lee1999LearningTP} is characterized by the inherent non-negativity of the latent representation of the data it provides. Ensuring non-negativity is computationally challenging, but provides unique advantages. For example, clustering in latent space is straightforward by assigning each sample or feature to the component with the highest loading. Furthermore, each cluster is easily interpretable due to the sparsity of the component loadings \cite{Brunet2004MetagenesAM}. In contrast, other latent representation methods such as Principal Component Analysis (PCA) or Independent Component Analysis (ICA) produce loadings with both positive and negative values. This can result in complex cancellations, making clustering and interpretation less intuitive. NMF has been adapted to satisfy domain-specific requirements through numerous extensions. To name a few:  Sparse NMF \cite{Hoyer2004NonnegativeMF} and Nonsmooth NMF \cite{PascualMontano2006NonsmoothNM} further improve the intrinsic sparsity of the component loadings; Graph Regularized NMF \cite{Cai2011GraphRN} aims to respect the intrinsic geometric structure of the data, leading to even more interpretable and meaningful results. Further contributing to the success of NMF in fields as diverse as bioinformatics, text mining, and image analysis, semi-NMF \cite{Ding2010ConvexAS} and non-negative tensor factorization (NTF) \cite{Cichocki2009FastLA} extend the applicability of NMF to signed data (i.e., with positive and negative values) and tensors of arbitrary order. \newline
 
Similarly, in survival analysis, leveraging meaningful NMF components to represent a large set of baseline covariates can help uncover covariate patterns that affect patient survival while capturing interdependencies among multiple covariates. By identifying such patterns, researchers can, for example, develop tailored treatment strategies for clusters of patients with one or another pattern based on their risk ratio on survival. This approach contrasts sharply with traditional models such as the Cox proportional hazards model, which assesses the risk ratio of individual covariates on survival but does not inherently reveal specific covariate patterns. However, we found that replacing the original data with their NMF latent representation in advanced survival models such as Coxnet \cite{Simon2011RegularizationPF} degrades the survival prediction performance in the majority of the datasets analyzed in this study, because the NMF analysis of baseline covariates does not account for the related survival information. To address the potential dissociation between survival prediction performance and clustering, a two-step approach is proposed in \cite{Li2024ConstructingAI}.  In the first step, a Coxnet approach is applied, incorporating elastic net regularization into the Cox proportional hazards model to address overfitting.  This approach effectively identifies baseline covariates, which, when considered collectively, are strongly predictive of survival. In the second step, NMF is applied to these identified covariates. However, there is no guarantee that the resulting clusters will only involve homogeneous survival outcomes. By incorporating Cox's partial likelihood into the NMF loss function, CoxNMF \cite{Huang2020LowRankRV} attempts to address the problem directly. Because the loss function relies on survival data for its computation, mapping the baseline covariates of new patients to the model's latent space—defined by learned covariate patterns—requires prior knowledge of their survival data. As a result, this method cannot be used to predict survival outcomes, which is a serious limitation. \newline
 
In this work, we present a novel approach: CoxNTF, which finds a non-negative latent representation of the baseline covariates that is closely associated with survival. Unlike CoxNMF, CoxNTF does not use a custom loss function. Instead, the baseline covariates are organized in a three-dimensional array. The third dimension consists of a finite set of time periods during which the expected event can be observed. For each time period, the covariates associated with a given sample are weighted according to the probability that the event is observed within that period. This probability is calculated as the difference between the restricted mean survival probabilities at the boundaries of the period, derived using the Coxnet approach. The resulting tensor undergoes an NTF decomposition. Instead of the original covariates, the NTF sample loadings are then used as input to train a new Coxnet model. Our results demonstrate that representing baseline covariates in the NTF latent space achieves comparable survival prediction performance to using the original covariates. Furthermore, this approach often outperforms the prediction accuracy—as evaluated using the concordance-index also known as the c-index \cite{Harrell1996}—obtained with the NMF latent representation. At the same time, similar to the functionality of NMF, the NTF representation allows for easy clustering. CoxNTF notably utilizes the Fast-HALS algorithm (Fast Hierarchical Alternating Least Squares) \cite{Cichocki2009FastLA}, recognized as a top-tier algorithm due to its exceptional efficiency and convergence performance \cite{Hou2024ConvergenceOA}. \newline

In summary, the CoxNTF approach demonstrates its computational effectiveness in clustering samples, while the easier to interpret covariate patterns can be used to predict survival in place of the original covariates using the same survival approach, such as Coxnet, with no loss of predictive power.

\section{Materials}

We analyzed 9 survival datasets:
\begin{itemize}
    \item "veterans\_lung\_cancer": Data from the Veterans’ Administration Lung Cancer Trial \cite{Kalbfleisch2002TheSA}. The  dataset has 137 samples and 6 features. The endpoint is death, which was observed in a majority of patients (93\%).
    \item "flchain": assay of serum free light chain \cite{Dispenzieri2012UseON}. The dataset has 7874 samples and 9 features. The endpoint is death, which was observed in a minority of patients (27\%).
    \item "whas500": Worcester Heart Attack Study dataset \cite{Hosmer2008AppliedSA}. The dataset has 500 samples and 14 features. The endpoint is death, which was observed in nearly half of the patients (43\%).
    \item "breast\_cancer": Breast cancer dataset \cite{Desmedt2007StrongTD}. The dataset has 198 samples and 80 features. The endpoint is the presence of distal metastases, which was observed in a minority of patients (25\%).
    \item "aids": AIDS Clinical Trial dataset \cite{Hosmer2008AppliedSA}. The dataset has 1,151 samples and 11 features. The event is the AIDS defining event (i.e. the suspected transmission-causing event), which was observed in a minority of patients (8\%).
    \item "gbsg2": German Breast Cancer Study Group 2 dataset \cite{Schumacher1994Randomized2X}. The dataset has 686 samples and 8 features. The endpoint is recurrence, and the time to recurrence is recurrence free survival, which observed in nearly half of the patients (43\%).
    \item "ds1": Churn Modelling dataset (\href{https://www.kaggle.com/datasets/shubh0799/churn-modelling}{churn-modelling}). The dataset has 10000 samples (employees of a bank) and 9 features (employee's attributes). The endpoint is an employee leaving the company, which was observed in a minority of employees (20\%).
    \item "ds2": Telco Customer Churn dataset (\href{https://www.kaggle.com/datasets/blastchar/telco-customer-churn/data}{telco-customer-churn/data}). The dataset has 7043 rows (customers of a company) and 17 features (customers's attributes). The endpoint is a customer leaving, which was observed in a minority of customers (27\%).
    \item "ds3": Customer Churn dataset (\href{https://www.kaggle.com/datasets/barun2104/telecom-churn}{telecom-churn}). The dataset has 3333 rows (customers of a telecom company) and 9 features (customers's attributes related to the services used). The endpoint is a customer leaving, which was observed in a minority of customers (14\%).
\end{itemize}
 
\section{Methods}
CoxNTF involves a two-step approach:
\begin{enumerate}
\item \textbf{NTF Model Training}: The approach begins by transforming the two-dimensional covariate matrix and survival data (time-to-event and censoring indicator) into a three-dimensional tensor representation. The third dimension consists of a finite set of time periods during which the expected event can be observed. For each time period, the covariates associated with a given sample are given a weight of 1 if the event is observed within that time period, and 0 otherwise. Additionally, inverse probability weighting is applied to account for censored observations. The tensor is then decomposed using Non-Negative Tensor Factorization (NTF), allowing for the discovery of key covariate structures and survival-related patterns.
\item \textbf{CoxNTF Model Training}: In the next stage, the Coxnet probability estimates are used to construct a weighted covariate tensor. As in the first step, the third dimension consists of a finite set of time periods during which the expected event can be observed. For each time period, the covariates associated with a given sample are weighted according to the probability that the event is observed within that period. This probability is calculated as the difference between the restricted mean survival probabilities at the boundaries of the period, derived using the Coxnet approach. The weighted covariate tensor is then projected onto a latent space obtained by the NTF decomposition from the first step, with the identified NTF features and temporal patterns remaining unchanged. The resulting meta-scores, which encapsulate critical survival-related features, serve as inputs to the final Coxnet model, called \textit{CoxNTF}.  As demonstrated in the Results section, this approach ensures effective survival-related classification while preserving the predictive efficiency of the Coxnet model.
\end{enumerate}

\begin{figure}[H]
    \centering
    \includegraphics[width=0.8\linewidth]{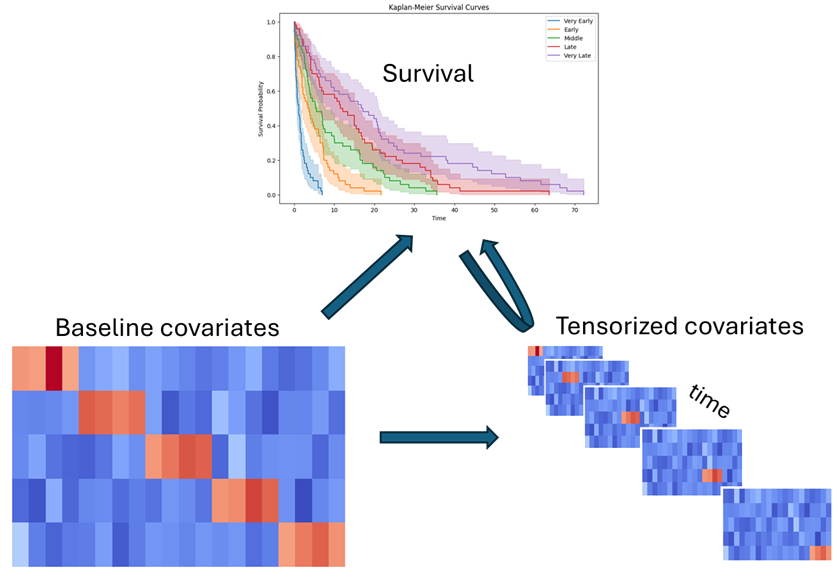}
    \caption{Interdependence between covariate matrix, tensor and survival}
    \label{fig:triangle}
\end{figure}

The interdependence among the three primary data objects—covariate matrix, survival information, and tensor representation—is depicted in Figure \ref{fig:triangle}. The feedback mechanism illustrated by the cycling arrow between survival and the tensor highlights the iterative refinement process: The tensor is initially constructed using baseline covariates and survival data, and subsequently, this structured representation contributes to an enhanced survival estimation, incorporating latent patterns and relationships within the data.

\subsection{NTF Model Training (algorithm \ref{alg:train_ntf})}\label{ntf_training_algorithm}

The NTF model is trained using a structured approach that incorporates survival data and covariate information while handling censored observations. \newline

\textbf{Input Data}:
 We define the input data as:
\renewcommand{\labelitemi}{\tiny$\bullet$}
\begin{itemize}
    \item A covariate matrix $\mathbf{X} \in \mathbb{R}_{+}^{n \times p}$, where $n$ is the number of observations, and $p$ is the number of covariates.
    \item A censoring indicator vector $\mathbf{c} \in \{0,1\}^n$ to distinguish observed survival times ($\mathbf{c}(i) = 1$) from censored ones ($\mathbf{c}(i) = 0$).
    \item A survival time vector $\mathbf{t} \in \mathbb{R}^n$, denoting the observed or censored times for each observation.
    \item A sequence of predefined survival periods $\mathbf{s} \in \mathbb{R}^{q+1}$, partitioning the time range into $q$ fixed intervals: $(\mathbf{s}(k-1), \mathbf{s}(k)]$. Typical 10, 25, 50, 75, and 90 percentiles of the distribution of  $\mathbf{t}$ are used to define these time periods. The need to have a reasonable and balanced sample size across time periods dictates the choice of percentiles. 
    \item The number of NTF components, denoted by $r$.
\end{itemize}

\textbf{Tensor Construction}:
 A three-dimensional tensor $\mathcal{X} \in \mathbb{R}_{+}^{n \times p \times q}$ is initialized with zeros. For each observation $i$, if the event was observed ($\mathbf{c}(i) = 1$), the corresponding covariate values are assigned to the appropriate survival period $k$:
\[
\mathcal{X}{}(i,:,k) \gets \mathbf{X}(i,:) \quad \text{if} \quad \mathbf{s}(k-1) < \mathbf{t}(i) \leq \mathbf{s}(k).
\]

\textbf{Inverse Probability of Censoring Weight (IPCW) Adjustment}:
To account for censoring, we use inverse probability weighting based on the Kaplan-Meier estimator, where censoring is considered the observed event. The tensor $\mathcal{X}$ is weighted by multiplying each row by the square root of its respective IPCW weight $ipcw_i$:
\[
\mathcal{X}{}(i,:,:) \gets \mathcal{X}{}(i,:,:) \times \sqrt{ipcw_i}.
\]
Taking the square root of \$ipcw\_i\$ ensures correct weighting in the quadratic loss function used by FAST-HALS.

\textbf{NTF Decomposition}:
 The weighted tensor $\mathcal{X}$ is factorized using NTF with $r$ components:
\[
\mathcal{X} \approx \sum_{1 \leq k \leq r} \mathbf{w}_k \circ \mathbf{h}_k \circ \mathbf{q}_k
\]
where the $\circ$ symbol denotes the outer product and:
\begin{itemize}
    \item  $\mathbf{W} = \left[ \mathbf{w}_1 , \mathbf{w}_2 , \dots , \mathbf{w}_r \right] \in \mathbb{R}_{+}^{n \times r}$ represents the meta-scores.
    \item $\mathbf{H} = \left[ \mathbf{h}_1 , \mathbf{h}_2 , \dots , \mathbf{h}_r \right] \in \mathbb{R}_{+}^{p \times r}$ represents patterns of covariates.
    \item $\mathbf{Q} = \left[ \mathbf{q}_1 , \mathbf{q}_2 , \dots , \mathbf{q}_r \right] \in \mathbb{R}_{+}^{q \times r}$ captures temporal survival patterns.

\end{itemize}

\textbf{Output}:
 The trained model returns the matrices $\mathbf{H}$ and $\mathbf{Q}$ which characterize survival across different covariate structures that are linked to temporal survival patterns.

\begin{algorithm}[H]
\caption{Estimation of CoxNTF patterns}\label{alg:train_ntf}
\begin{algorithmic}[1]
\Require $\mathbf{X} \in \mathbb{R}_{+}^{n \times p}$, $\mathbf{c} \in \{0,1\}^n$, $\mathbf{t} \in \mathbb{R}^n$, $\mathbf{s} \in \mathbb{R}^{q+1}$,  NTF rank $r$
\Ensure Matrices $\mathbf{H} \in \mathbb{R}_{+}^{p \times r}$ and $\mathbf{Q} \in \mathbb{R}_{+}^{q \times r}$

\State Initialize tensor $\mathcal{X} \in \mathbb{R}_{+}^{n \times p \times q}$ with zeros
\For{$i = 1$ to $n$}
    \If{$\mathbf{c}(i) = 1$}
        \For{$k = 1$ to $q$}
            \If{$\mathbf{s}(k-1) < \mathbf{t}(i) \leq \mathbf{s}(k)$}
                \State $\mathcal{X}(i,:,k) \gets \mathbf{X}(i,:)$
            \EndIf
        \EndFor
    \EndIf
\EndFor

\State Compute inverse probability of censoring weights $\{ipcw_i\}$ using Kaplan-Meier estimator
\For{$i = 1$ to $n$}
    \State $\mathcal{X}(i,:,:) \gets \mathcal{X}(i,:,:) \times \sqrt{ipcw_i}$
\EndFor

\State Perform Nonnegative Tensor Factorization (NTF)
\State Approximate $\mathcal{X}$ using $r$ components:
\[
\mathcal{X} \approx \sum_{1 \leq k \leq r} \mathbf{w}_k \circ \mathbf{h}_k \circ \mathbf{q}_k
\]

\State Extract matrices $\mathbf{H}$ and $\mathbf{Q}$
\State \Return $\mathbf{H}$, $\mathbf{Q}$
\end{algorithmic}
\end{algorithm}

\subsection{CoxNTF Model Training (algorithm \ref{alg:train_coxntf})}

To extract meaningful meta-scores for survival prediction and classification, the CoxNTF model integrates Coxnet survival probabilities with NTF covariate structures and temporal survival patterns found in the first stage (section \ref{ntf_training_algorithm}).
\newline

\textbf{Input Data}:
 The same input data as in the first stage are required for the integration of the Coxnet model as well as the estimated NTF matrices $\mathbf{H}$ and $\mathbf{Q}$.
\newline

\textbf{Coxnet Probability Estimation}:
 A Coxnet survival model is trained based on $\mathbf{X}, \mathbf{c},$ and $\mathbf{t}$.
 Using this model, we estimate the probability \text{$\mathbb{P}(\mathbf{X}(i,:),k)$} that an event occurs within survival period $k$.
\newline

\textbf{Tensor Construction}:
A three-dimensional tensor $\mathcal{X} \in \mathbb{R}^{n \times p \times q}$ is initialized as zeros.
For each observation $i$ and each survival period $k$, covariate values are weighted by the square-root of Coxnet event probability within period $k$:
    \[
    \mathcal{X}(i,:,k) \gets \mathbb{P}(\mathbf{X}(i,:),k)^{1/2} \times \mathbf{X}(i,:)
    \]

\textbf{NTF mapping}:
 The NTF decomposition is applied to $\mathcal{X}$ with fixed basis vectors $\mathbf{H}$ and $\mathbf{Q}$.
 This decomposition yields the matrix of meta-scores:
    \[
    \mathbf{W} = \left[ \mathbf{w}_1 , \mathbf{w}_2 , \dots , \mathbf{w}_r \right] \in \mathbb{R}^{n \times r}
    \]

\textbf{CoxNTF Model Estimation}:
 The matrix $\mathbf{W}$ is used in place of the original covariates to estimate a Coxnet model, referred to as \textit{CoxNTF model}, i.e. a Coxnet model is trained on $\mathbf{W}, \mathbf{c},$ and $\mathbf{t}$.
\newline

\textbf{Output}:
 The final CoxNTF model returns the matrix of meta-scores $\mathbf{W}$ summarizing survival-relevant factors and the CoxNTF model parameters derived from $\mathbf{W}$.

\begin{algorithm}[H]
\caption{CoxNTF Model Estimation}\label{alg:train_coxntf}
\begin{algorithmic}[1]
\Require Covariate matrix $\mathbf{X} \in \mathbb{R}_{+}^{n \times p}$, Censoring indicator vector $\mathbf{c} \in \{0,1\}^n$, Survival time vector $\mathbf{t} \in \mathbb{R}^n$, Fixed survival periods $\mathbf{s} \in \mathbb{R}^{q+1}$, Number of NTF components $r$, Covariate pattern matrix $\mathbf{H} \in \mathbb{R}_{+}^{p \times r}$, Temporal survival pattern matrix $\mathbf{Q} \in \mathbb{R}_{+}^{q \times r}$.
\Ensure Meta-score matrix $\mathbf{W} \in \mathbb{R}_{+}^{n \times r}$ and CoxNTF model parameters.

\State Train Coxnet survival model using $\mathbf{X}, \mathbf{c}, \mathbf{t}$
\For{each observation $i$}
    \For{each survival period $k$}
        \State Estimate $\mathbb{P}(\mathbf{X}(i,:), k)$ using Coxnet model
    \EndFor
\EndFor

\State Initialize tensor $\mathcal{X} \in \mathbb{R}^{n \times p \times q}$ with zeros
\For{each observation $i$}
    \For{each survival period $k$}
        \State $\mathcal{X}(i,:,k) \gets \mathbb{P}(\mathbf{X}(i,:), k)^{1/2} \times \mathbf{X}(i,:)$
    \EndFor
\EndFor

\State Apply NTF decomposition to $\mathcal{X}$ using fixed basis vectors $\mathbf{H}, \mathbf{Q}$
\State Extract meta-score matrix $\mathbf{W} \in \mathbb{R}^{n \times r}$

\State Train CoxNTF model using $\mathbf{W}, \mathbf{c}, \mathbf{t}$
\State Return $\mathbf{W}$ and CoxNTF model parameters
\end{algorithmic}
\end{algorithm}

\subsection{NTF Rank determination}
To determine the appropriate number of NTF components (or \textit{rank}) during NTF model training, the dataset is randomly divided into \textit{training} and \textit{validation} subsets. The optimal rank is selected based on the CoxNTF model, identifying the value that yields the highest c-index in the \textit{validation} subset. To improve reliability, this random partitioning process can be repeated multiple times, ensuring a stable mean estimate.

\subsection{Data and Code availability}
The first six datasets are easily accessible via the Datasets API of the Python package \href{https://scikit-survival.readthedocs.io/en/stable/api/index.html}{scikit-survival}. The last three datasets are made available in the "parquet" format  in our github repository \href{https://github.com/Advestis}{R\&D Forvis Mazars · GitHub}. Python code (modules and notebooks) to reproduce the results is also available.

\section{Results}

We compared three approaches: Coxnet, CoxNMF, and CoxNTF. In the Coxnet approach, the original features are used, whereas in the CoxNMF and CoxNTF approaches, Coxnet takes the NMF and NTF latent factors as input. The following criteria were used:
\begin{itemize}
    \item The concordance index (c-index) for right-censored data based on inverse probability of censoring weights \cite{Uno2011OnTC}.
    \item The number of features retained by the Coxnet model, where the features are either the original dataset features or the latent factors found by either NMF or NTF.
\end{itemize}

Each dataset was randomly partitioned into training, validation, and test datasets to train, evaluate, and test the best model for each approach. Random partitions were repeated 30 times and the criteria were averaged. The results are presented in table \ref{tab:comparison}. \newline

Our results consistently show that the Coxnet and CoxNTF approaches achieve very similar c-indexes, always superior to CoxNMF, sometimes significantly so for the "flchain" dataset (c-index=0.77 for CoxNTF versus 0.68 for CoxNMF). In addition, the Coxnet model retains a substantially lower average number of features when NTF latent factors are used instead of NMF latent factors (up to 45\% less). Taken together, these results are consistent with our expectations: NTF latent factors, in contrast to NMF latent factors, provide a survival-consistent classification with predictive accuracy similar to the Coxnet model. \newline

\begin{figure*}
    \centering
    \includegraphics[width=0.75\linewidth]{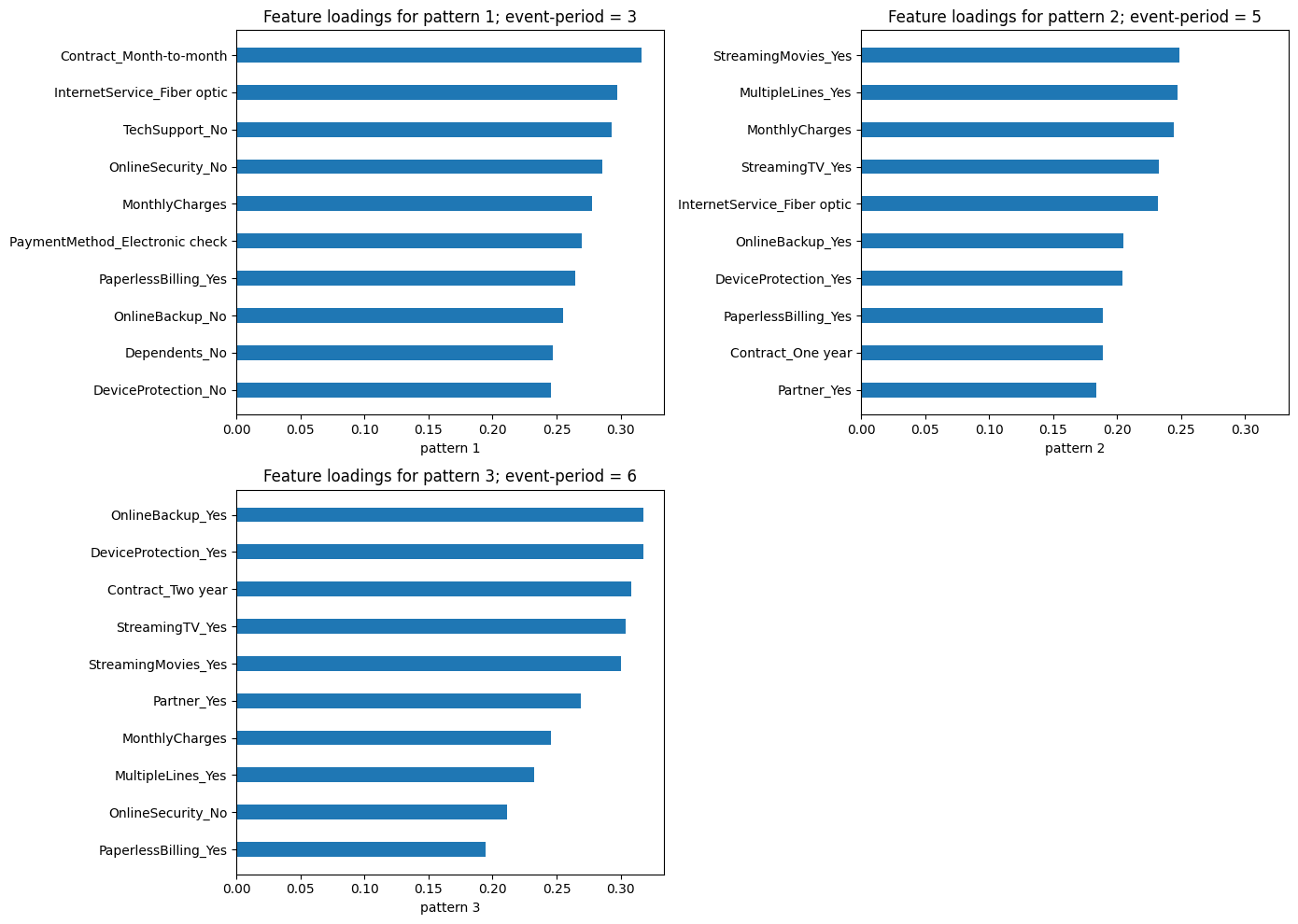}
    \caption{NTF patterns found in the "ds2" analysis}
    \label{fig:NTF-patterns}
\end{figure*}

\begin{table*}
    \centering
    \caption{Comparison of different survival approaches across datasets}
    \label{tab:comparison}
    \resizebox{\textwidth}{!}{
    \begin{tabular}{|l|c|c|c|c|c|c|}
    \hline
        \textbf{Approach} & \multicolumn{2}{c|}{\textbf{veterans\_lung\_cancer (137, 6)}} & \multicolumn{2}{c|}{\textbf{flchain (7874, 9)}} & \multicolumn{2}{c|}{\textbf{whas500 (500, 14)}} \\ \hline
        ~ & Average \# features & c-index& Average \# features & c-index& Average \# features & c-index\\ \hline
        \textbf{NTF} & 3.7 & 0.71 & 1.5 & 0.77 & 2.7 & 0.75 \\ \hline
        \textbf{NMF} & 5.4 & 0.65 & 3.2 & 0.68 & 4.3 & 0.66 \\ \hline
        \textbf{COX} & 4.6 & 0.70& 7.8 & 0.78 & 9.5 & 0.76 \\ \hline
        ~ & ~ & ~ & ~ & ~ & ~ & ~ \\ \hline
        ~ &  \multicolumn{2}{c|}{\textbf{breast\_cancer (198, 80)}} &  \multicolumn{2}{c|}{\textbf{gbsg2 (686, 8)}} &  \multicolumn{2}{c|}{\textbf{aids (1151, 11)}} \\ \hline
           ~ & Average \# features & c-index& Average \# features & c-index& Average \# features & c-index\\ \hline
        \textbf{NTF} & 2.2 & 0.66 & 2.8 & 0.66 & 3.2 & 0.70\\ \hline
        \textbf{NMF} & 3.7 & 0.62 & 5.1 & 0.64 & 4.9 & 0.68 \\ \hline
        \textbf{COX} & 29 & 0.66 & 15.3 & 0.66 & 12.2 & 0.70\\ \hline
        ~ & ~ & ~ & ~ & ~ & ~ & ~ \\ \hline
         ~ &  \multicolumn{2}{c|}{\textbf{ds1 (10000, 9)}} &  \multicolumn{2}{c|}{\textbf{ds2 (7043, 17)}} &  \multicolumn{2}{c|}{\textbf{ds3 (3333, 9)}} \\ \hline
           ~ & Average \# features & c-index& Average \# features & c-index& Average \# features & c-index\\ \hline
        \textbf{NTF} & 2.9 & 0.72 & 3.8 & 0.78 & 3.9 & 0.75 \\ \hline
        \textbf{NMF} & 4.5 & 0.67 & 4.9 & 0.74 & 4.5 & 0.71 \\ \hline
        \textbf{COX} & 9.8 & 0.71 & 17.5 & 0.78 & 12.7 & 0.73 \\ \hline
    \end{tabular}
    }
\end{table*}
For illustrative purposes, the NTF patterns found in the analysis of the "ds2" dataset (\href{https://www.kaggle.com/datasets/blastchar/telco-customer-churn/data}{telco-customer-churn/data}) are shown in Figure \ref{fig:NTF-patterns}. The first pattern is associated with customers leaving earlier (event period = 3). In particular, a month-to-month contract is a major feature contributing to early departure. The other two patterns are associated with customers leaving later (event-period  $\geq 5$). Unsurprisingly, these customers are on one or two year contracts and use device protection or online backup services, in contrast to those following the first pattern.

\begin{figure*}
    \centering
    \includegraphics[width=0.75\linewidth]{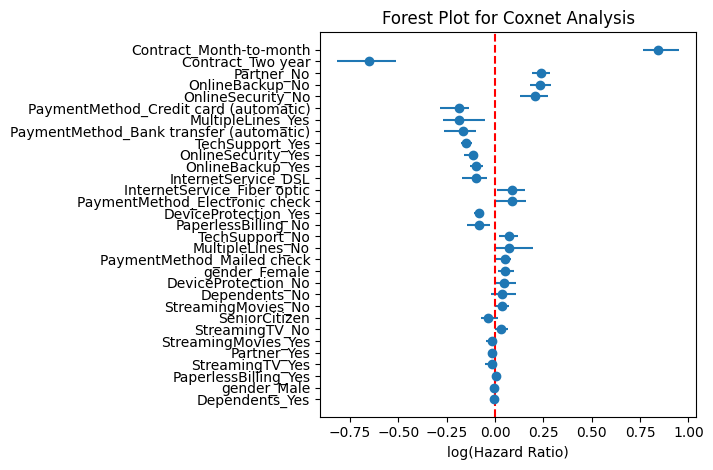}
    \caption{Coxnet forest plot}
    \label{fig:forest-plot}
\end{figure*}

The Coxnet forest plot of hazard ratios is shown in Figure \ref{fig:forest-plot}. Month-to-month contracts and the absence of services such as online backup of security carry a positive hazard ratio, consistent with the first NTF pattern. The other prominent features seen in the other two NTF patterns are the two-year contract and the use of the above services, which carry negative hazard ratios. \newline

Coxnet requires categorical variables to be transformed into numerical binary representations. To achieve this, dummy coding generates a set of binary variables that correspond to each category within a categorical variable. Due to perfect collinearity among month-to-month, one-year, and two-year categories, the elastic net model assigns a value of zero to the one-year contract. While this exclusion is somewhat arbitrary—any of the three categories could have been set to zero—an expanded estimate of the one-year contract could be obtained by calculating the inverse of the geometric mean of the hazard ratios for the other two contract types. However, this approach may be affected by the potential entanglement of binary variables, independent of their parent categorical variable, of which the model is unaware, resulting in some categories being close to zero rather than being completely excluded. For instance, gender remains fully represented in the forest plot, with the female category displaying a positive hazard ratio, while the male category is positioned close to zero. \newline

In this example, the elastic net does not consistently identify complete redundancy among binary variables because its performance is influenced by the interdependences among all binary variables and the applied penalty. As a result, interpreting a Coxnet forest plot can be challenging. In contrast, CoxNTF effectively handles collinearities without arbitrarily eliminating certain categories, allowing the identification of consistent relationships among them.

\section{Discussion}

CoxNTF is a semi-supervised approach that integrates both supervised and unsupervised learning. It first builds a tensor representation of covariate matrices using survival data. It then applies Nonnegative Tensor Factorization (NTF) to this tensor in an unsupervised manner, identifying a limited number of covariate patterns strongly linked to survival. These patterns can replace the original covariates in survival prediction without loss of accuracy and facilitate patient classification into survival-associated categories. \newline

Unlike standard Cox models, which assess risk factors for individual covariates, CoxNTF uncovers meaningful patterns among multiple covariates. For the practitioner, this reduces ambiguity when evaluating a patient's survival risk, offering clearer insight into covariate interactions.\newline

The CoxNTF framework can be readily adapted to incorporate time-varying covariates with a minor modification to algorithm \ref{alg:train_coxntf}. Specifically: (i) The weighting procedure should employ an extended Cox model capable of handling time-varying covariates. (ii) Instead of assigning baseline-weighted covariates across all time periods, the tensor should be constructed by weighting covariate values according to their specific time points. Additionally, the temporal segmentation may need to be revised to align with the timing of covariate changes. Although none of the examples in this study utilize time-varying covariates, exploring this extension is planned for future work. It is worth emphasizing that the Cox model with time-varying covariates is particularly well-suited for retrospective analysis—revealing how temporal changes in covariates influence event hazards. However, its utility for prediction is limited unless a strategy is in place to manage future covariate values, which are typically unknown at the time of prediction. \newline

This work has certain limitations. More research is required to evaluate the impact of selecting the number and percentile levels used to define time periods on the performance of CoxNTF. Additionally, CoxNTF is not a standalone survival prediction method, as it depends on existing models for its predictions. Importantly, its purpose is not to enhance the survival prediction accuracy of the models it interacts with.\newline

Some directions for future work include leveraging advanced deep learning-based Cox models as foundational survival models, enhancing CoxNTF's role as a complementary classification tool to augment state-of-the-art survival analysis methods with pattern-based interpretation.


\begin{thebibliography}{18}
\providecommand{\natexlab}[1]{#1}
\providecommand{\url}[1]{\texttt{#1}}
\providecommand{\urlprefix}{URL }
\expandafter\ifx\csname urlstyle\endcsname\relax
  \providecommand{\doi}[1]{doi:\discretionary{}{}{}#1}\else
  \providecommand{\doi}{doi:\discretionary{}{}{}\begingroup \urlstyle{rm}\Url}\fi

\bibitem[{Brunet et~al.(2004)Brunet, Tamayo, Golub, and Mesirov}]{Brunet2004MetagenesAM}
Brunet, J.-P.; Tamayo, P.; Golub, T.~R.; and Mesirov, J.~P. 2004.
\newblock Metagenes and molecular pattern discovery using matrix factorization.
\newblock \emph{Proceedings of the National Academy of Sciences} 101(12): 4164--4169.
\newblock \doi{10.1073/pnas.0308531101}.

\bibitem[{Cai et~al.(2011)Cai, He, Han, and Huang}]{Cai2011GraphRN}
Cai, D.; He, X.; Han, J.; and Huang, T.~S. 2011.
\newblock Graph Regularized Nonnegative Matrix Factorization for Data Representation.
\newblock \emph{IEEE Transactions on Pattern Analysis and Machine Intelligence} 33(8): 1548--1560.
\newblock \doi{10.1109/TPAMI.2010.231}.

\bibitem[{Cichocki and Phan(2009)}]{Cichocki2009FastLA}
Cichocki, A.; and Phan, A.-H. 2009.
\newblock Fast Local Algorithms for Large Scale Nonnegative Matrix and Tensor Factorizations.
\newblock \emph{IEICE Transactions on Fundamentals of Electronics, Communications and Computer Sciences} E92.A(3): 708--721.
\newblock \doi{10.1587/transfun.E92.A.708}.

\bibitem[{Desmedt et~al.(2007)Desmedt, Piette, Loi, Wang, Lallemand, Haibe-Kains, Viale, Delorenzi, Zhang, D'assignies, Bergh, Lidereau, Ellis, Harris, Klijn, Foekens, Cardoso, Piccart, Buyse, and Sotiriou}]{Desmedt2007StrongTD}
Desmedt, C.; Piette, F.; Loi, S.; Wang, Y.; Lallemand, F.; Haibe-Kains, B.; Viale, G.; Delorenzi, M.; Zhang, Y.; D'assignies, M.~S.; Bergh, J.; Lidereau, R.; Ellis, P.~A.; Harris, A.~L.; Klijn, J. G.~M.; Foekens, J.~A.; Cardoso, F.; Piccart, M.~J.; Buyse, M.~E.; and Sotiriou, C. 2007.
\newblock Strong Time Dependence of the 76-Gene Prognostic Signature for Node-Negative Breast Cancer Patients in the TRANSBIG Multicenter Independent Validation Series.
\newblock \emph{Clinical Cancer Research} 13: 3207 -- 3214.
\newblock \urlprefix\url{https://aacrjournals.org/clincancerres/article/13/11/3207/193398}.

\bibitem[{Ding, Li, and Jordan(2010)}]{Ding2010ConvexAS}
Ding, C.~H.; Li, T.; and Jordan, M.~I. 2010.
\newblock Convex and Semi-Nonnegative Matrix Factorizations.
\newblock \emph{IEEE Transactions on Pattern Analysis and Machine Intelligence} 32(1): 45--55.
\newblock \doi{10.1109/TPAMI.2008.277}.

\bibitem[{Dispenzieri et~al.(2012)Dispenzieri, Katzmann, Kyle, Larson, Therneau, Colby, Clark, Mead, Kumar, Melton, and Rajkumar}]{Dispenzieri2012UseON}
Dispenzieri, A.; Katzmann, J.~A.; Kyle, R.~A.; Larson, D.~R.; Therneau, T.~M.; Colby, C.~L.; Clark, R.~J.; Mead, G.~P.; Kumar, S.~K.; Melton, L.~J.; and Rajkumar, S.~V. 2012.
\newblock Use of nonclonal serum immunoglobulin free light chains to predict overall survival in the general population.
\newblock \emph{Mayo Clinic proceedings} 87 6: 517--23.
\newblock \urlprefix\url{https://pubmed.ncbi.nlm.nih.gov/22677072/}.

\bibitem[{Harrell, Lee, and Mark(1996)}]{Harrell1996}
Harrell, F.~E.; Lee, K.~L.; and Mark, D.~B. 1996.
\newblock Multivariable prognostic models: Issues in developing models, evaluating assumptions and adequacy, and measuring and reducing errors.
\newblock \emph{Statistics in Medicine} 15(4): 361--387.
\newblock \doi{10.1002/(SICI)1097-0258(19960229)15:4<361::AID-SIM168>3.0.CO;2-4}.

\bibitem[{Hosmer, Lemeshow, and May(2008)}]{Hosmer2008AppliedSA}
Hosmer, D.~W.; Lemeshow, S.; and May, S.~J. 2008.
\newblock Applied Survival Analysis: Regression Modeling of Time-to-Event Data.
\newblock \urlprefix\url{https://academic.oup.com/biometrics/article-abstract/65/2/671/7331827}.

\bibitem[{Hou, Chu, and Liao(2024)}]{Hou2024ConvergenceOA}
Hou, L.; Chu, D.; and Liao, L.-Z. 2024.
\newblock Convergence of a Fast Hierarchical Alternating Least Squares Algorithm for Nonnegative Matrix Factorization.
\newblock \emph{IEEE Transactions on Knowledge and Data Engineering} 36(1): 77--89.
\newblock \doi{10.1109/TKDE.2023.3279369}.

\bibitem[{Hoyer(2004)}]{Hoyer2004NonnegativeMF}
Hoyer, P.~O. 2004.
\newblock Non-negative Matrix Factorization with Sparseness Constraints.
\newblock \emph{Journal of Machine Learning Research} 5: 1457--1469.
\newblock \urlprefix\url{https://jmlr.org/papers/v5/hoyer04a.html}.

\bibitem[{Huang et~al.(2020)Huang, Salama, Shao, Zhang, and Huang}]{Huang2020LowRankRV}
Huang, Z.; Salama, P.; Shao, W.; Zhang, J.; and Huang, K. 2020.
\newblock Low-Rank Reorganization via Proportional Hazards Non-negative Matrix Factorization Unveils Survival Associated Gene Clusters.
\newblock \urlprefix\url{https://arxiv.org/abs/2008.03776}.

\bibitem[{Kalbfleisch and Prentice(2002)}]{Kalbfleisch2002TheSA}
Kalbfleisch, J.~D.; and Prentice, R.~L. 2002.
\newblock The Statistical Analysis of Failure Time Data: Kalbfleisch/The Statistical.
\newblock \urlprefix\url{https://onlinelibrary.wiley.com/doi/book/10.1002/9781118032985}.

\bibitem[{Lee and Seung(1999)}]{Lee1999LearningTP}
Lee, D.~D.; and Seung, H.~S. 1999.
\newblock Learning the parts of objects by non-negative matrix factorization.
\newblock \emph{Nature} 401(6755): 788--791.
\newblock \doi{10.1038/44565}.

\bibitem[{Li et~al.(2024)Li, Meng, Xiao, Du, Jiang, and Liu}]{Li2024ConstructingAI}
Li, Z.; Meng, Z.; Xiao, L.; Du, J.; Jiang, D.; and Liu, B. 2024.
\newblock Constructing and identifying an eighteen-gene tumor microenvironment prognostic model for non-small cell lung cancer.
\newblock \emph{World Journal of Surgical Oncology} 22: 319.
\newblock \doi{10.1186/s12957-024-03588-y}.

\bibitem[{Pascual-Montano et~al.(2006)Pascual-Montano, Carazo, Kochi, Lehmann, and Pascual-Marqui}]{PascualMontano2006NonsmoothNM}
Pascual-Montano, A.; Carazo, J.; Kochi, K.; Lehmann, D.; and Pascual-Marqui, R. 2006.
\newblock Nonsmooth nonnegative matrix factorization (nsNMF).
\newblock \emph{IEEE Transactions on Pattern Analysis and Machine Intelligence} 28(3): 403--415.
\newblock \doi{10.1109/TPAMI.2006.60}.

\bibitem[{Schumacher et~al.(1994)Schumacher, Bastert, Bojar, H{\"u}bner, Olschewski, Sauerbrei, Schmoor, Beyerle, Neumann, and Hf}]{Schumacher1994Randomized2X}
Schumacher, M.; Bastert, G.; Bojar, H.; H{\"u}bner, K.; Olschewski, M.; Sauerbrei, W.; Schmoor, C.; Beyerle, C.; Neumann, R. L.~A.; and Hf, R. 1994.
\newblock Randomized 2 x 2 trial evaluating hormonal treatment and the duration of chemotherapy in node-positive breast cancer patients. German Breast Cancer Study Group.
\newblock \emph{Journal of clinical oncology : official journal of the American Society of Clinical Oncology} 12 10: 2086--93.
\newblock \urlprefix\url{https://pubmed.ncbi.nlm.nih.gov/7931478/}.

\bibitem[{Simon et~al.(2011)Simon, Friedman, Hastie, and Tibshirani}]{Simon2011RegularizationPF}
Simon, N.; Friedman, J.; Hastie, T.; and Tibshirani, R. 2011.
\newblock Regularization Paths for Cox's Proportional Hazards Model via Coordinate Descent.
\newblock \emph{Journal of Statistical Software} 39(5): 1--13.
\newblock \doi{10.18637/jss.v039.i05}.

\bibitem[{Uno et~al.(2011)Uno, Cai, Pencina, D’Agostino, and Wei}]{Uno2011OnTC}
Uno, H.; Cai, T.; Pencina, M.~J.; D’Agostino, R.~B.; and Wei, L.-J. 2011.
\newblock On the C‐statistics for evaluating overall adequacy of risk prediction procedures with censored survival data.
\newblock \emph{Statistics in Medicine} 30.
\newblock \urlprefix\url{https://onlinelibrary.wiley.com/doi/10.1002/sim.4154}.

\end{thebibliography}

\end{document}